\documentclass[letterpaper]{article} % DO NOT CHANGE THIS
\usepackage{aaai2026}  % DO NOT CHANGE THIS
\usepackage{times}  % DO NOT CHANGE THIS
\usepackage{helvet}  % DO NOT CHANGE THIS
\usepackage{courier}  % DO NOT CHANGE THIS
\usepackage[hyphens]{url}  % DO NOT CHANGE THIS
\usepackage{graphicx} % DO NOT CHANGE THIS
\usepackage{natbib}  % DO NOT CHANGE THIS AND DO NOT ADD ANY OPTIONS TO IT
\usepackage{caption} % DO NOT CHANGE THIS AND DO NOT ADD ANY OPTIONS TO IT
\usepackage{algorithm}
\usepackage{algorithmic}
\usepackage[most]{tcolorbox}
\usepackage[table]{xcolor}

\usepackage[switch]{lineno}
\usepackage{multirow}
\usepackage{amsmath}
\usepackage{amssymb}
\usepackage{booktabs}
\usepackage{subcaption}
\usepackage{diagbox}
\usepackage{makecell}
\tcbuselibrary{breakable}    % Enables multi-page or multi-column breaking

\usepackage{soul}
\sethlcolor{yellow}

\usepackage{newfloat}
\usepackage{listings}
\DeclareCaptionStyle{ruled}{labelfont=normalfont,labelsep=colon,strut=off} % DO NOT CHANGE THIS
\floatstyle{ruled}
\newfloat{listing}{tb}{lst}{}
\floatname{listing}{Listing}
\title{Format Matters: The Robustness of Multimodal LLMs in Reviewing Evidence from Tables and Charts}

\author{
Xanh Ho,$^1$ 
Yun-Ang Wu,$^{2*}$ 
Sunisth Kumar,$^3$ \\
Florian Boudin,$^4$ 
Atsuhiro Takasu,$^1$ 
Akiko Aizawa$^{1,3}$ 
}
\affiliations{
$^1$National Institute of Informatics, Japan \hspace{1cm} 
$^2$National Taiwan University \\
$^3$The University of Tokyo, Japan \hspace{1cm}
$^4$Inria, LS2N, Nantes Université, France \\ 
{\tt \{xanh, takasu, aizawa\}@nii.ac.jp} \hspace{1cm}
{\tt r11944072@csie.ntu.edu.tw} \\
{\tt sunisth@g.ecc.u-tokyo.ac.jp} \hspace{1cm}
{\tt florian.boudin@univ-nantes.fr} 
}

\usepackage{bibentry}
\begin{document}

\maketitle

\begingroup\def\thefootnote{*}\footnotetext{Research conducted during internship at NII, Japan.}\endgroup

\begin{abstract}

% Background & Motivations
With the growing number of submitted scientific papers, there is an increasing demand for systems that can assist reviewers in evaluating research claims. Experimental results are a core component of scientific work, often presented in varying formats such as tables or charts. Understanding how robust current multimodal large language models (multimodal LLMs) are at verifying scientific claims across different evidence formats remains an important and underexplored challenge.
In this paper, we design and conduct a series of experiments to assess the ability of multimodal LLMs to verify scientific claims using both tables and charts as evidence. To enable this evaluation, we adapt two existing datasets of scientific papers by incorporating annotations and structures necessary for a multimodal claim verification task. 
Using this adapted dataset, we evaluate 12 multimodal LLMs and find that current models perform better with table-based evidence while struggling with chart-based evidence.
We further conduct human evaluations and observe that humans maintain strong performance across both formats, unlike the models. 
Our analysis also reveals that smaller multimodal LLMs (under 8B) show weak correlation in performance between table-based and chart-based tasks, indicating limited cross-modal generalization. 
These findings highlight a critical gap in current models' multimodal reasoning capabilities. We suggest that future multimodal LLMs should place greater emphasis on improving chart understanding to better support scientific claim verification.

% Results % Findings 

\end{abstract}

\begin{links}
    \link{Code}{https://github.com/Alab-NII/tables-vs-charts}
    % \link{Datasets}{https://github.com/Alab-NII/tables-vs-charts}
    % \link{Extended version}{https://aaai.org/example/extended-version}
\end{links}

\section{Introduction}

% Background - claim verification & its limitation: lack of dataset with the same evidence across formats
Scientific claim verification requires models to determine whether a given claim is supported or not, based on the provided evidence.
This evidence can take the form of text, tables, or charts. 
In recent years, several datasets have been introduced for this task, including SciFact~\cite{wadden-etal-2020-fact}, HealthVer~\cite{sarrouti-etal-2021-evidence-based}, SciTab~\cite{lu-etal-2023-scitab}, and MuSciClaims~\cite{lal2025musciclaims}. 
Depending on the dataset, the evidence may be textual (e.g., SciFact), tabular (e.g., SciTab), or visual (e.g., MuSciClaims).
However, existing datasets typically represent evidence in only a single format (details in Section~\ref{sec:related_work}). 
While the recent SciVer dataset~\cite{wang-etal-2025-sciver} incorporates multiple evidence modalities such as textual, tabular, and visual, it still presents tables in visual rather than text-based table format. 
Moreover, the different types of evidence in SciVer provide complementary information rather than expressing the same content across modalities.

% Motivations
In the era of generative AI and large language models (LLMs), researchers can now produce research papers more efficiently with support from AI agents~\cite{press2024citeme,si2025can}, leading to a significant increase in the number of submissions.
As a result, automated peer review systems and tools that assist human reviewers are becoming increasingly important.
A critical part of many research papers is the experimental results section, which presents key findings that support the main claims. 
These results are typically presented in tables or charts, depending on the author's preference. 
Understanding how multimodal LLMs handle different evidence formats is essential, especially as they are integrated into review systems. 
If LLMs perform well on one format but poorly on another, this could lead to biased or incomplete evaluations. 
Therefore, assessing their robustness across formats is vital for building reliable and generalizable AI-assisted review systems. 
However, it remains unclear whether current multimodal LLMs can consistently verify claims across different evidence formats. We argue that an effective review system must evaluate claims accurately, regardless of how the supporting evidence is presented.

\begin{figure*}[h]
    \centering
    \includegraphics[width=0.8\textwidth]{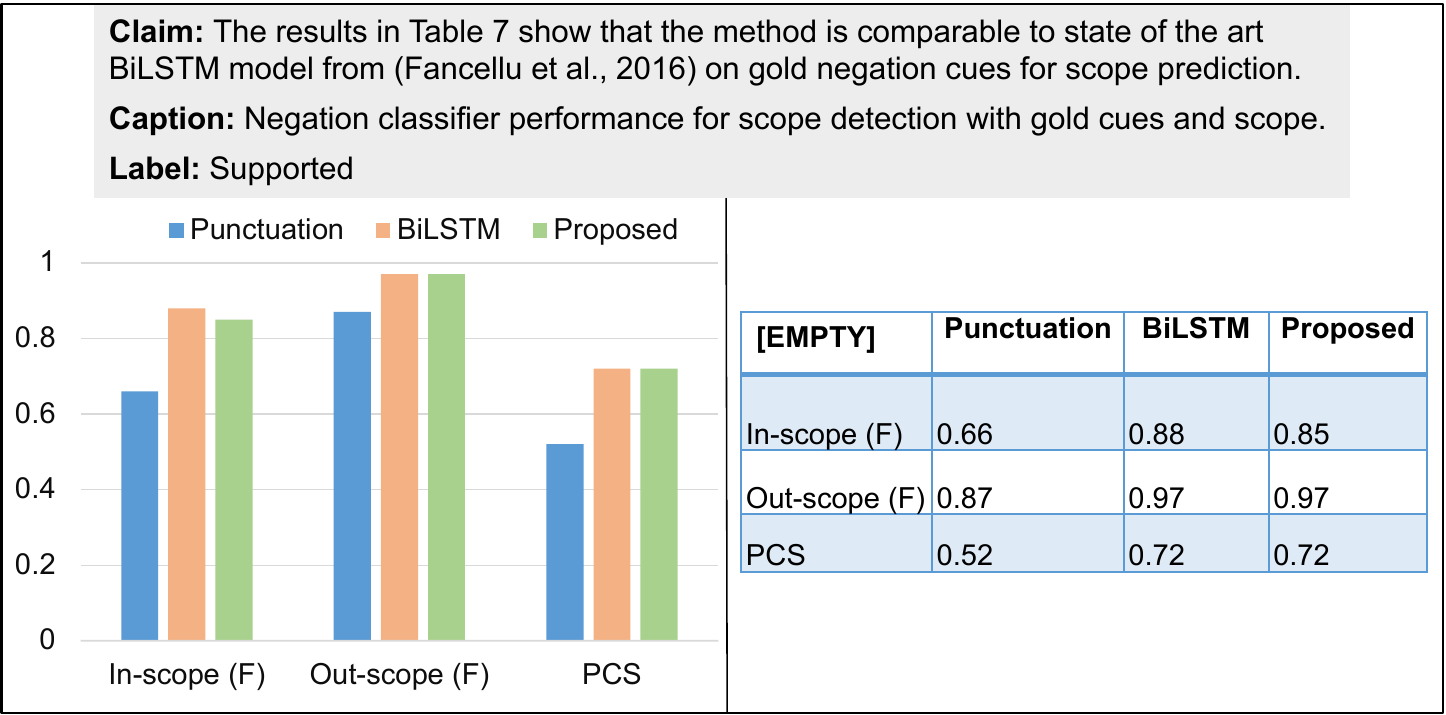}
    \caption{An example of the claim verification task in our experiment, featuring both types of evidence: table and chart formats that represent the same information.
    It is noted that the original example is from the SciTab dataset. 
    To ensure a fair comparison between table- and chart-evidence formats, we modified claims by replacing references to tables with figures (e.g., ``Table 7'' → ``Figure 7'') to make them compatible with chart-format evidence.
    }
    \label{fig:example}
\end{figure*}

% Another motivation - chart to code & its limitation: does not investigate the affect of the format to the main task --- but focus solely on converting 
From a different task perspective, Chart2Code and Data2Chart are tasks that focus on converting between charts and the code used to generate them, or on transforming tabular data into visual charts.
Several datasets have been developed for these tasks, such as MatPlotBench~\cite{yang-etal-2024-matplotagent}, Plot2Code~\cite{wu-etal-2025-plot2code} , and ChartMimic~\cite{yang2025chartmimic}. 
These datasets typically provide paired examples of code and the corresponding chart, with an emphasis on the transformation between these two formats.
While our research also involves both data formats, specifically charts and tables (assuming that tables can be derived from the underlying code), our focus is different. 
We treat these formats as distinct types of evidence for a downstream task, claim verification, rather than for format conversion.

% What will we do? 
In this paper, we aim to investigate the robustness of current multimodal LLMs in performing claim verification across different evidence formats. 
Specifically, we consider both tables and charts as representations of the evidence data. 
We found that existing datasets do not fully meet the requirements of our task, as they typically lack aligned table and chart evidence that convey the same underlying information. To address this gap, we extend two existing datasets, namely SciTabAlign~\cite{ho-etal-2025-table} and ChartMimic~\cite{yang2025chartmimic}, to construct a new dataset tailored to our claim verification setting. 
In particular, we curate instances that consist of a scientific claim, a table serving as evidence, and a corresponding chart that represents the same data as the table.
This alignment enables a systematic evaluation of multimodal LLM performance in verifying claims using different, yet equivalent, evidence formats.
Figure~\ref{fig:example} illustrates an example of the claim verification task used in our experiments.

Using the extended versions of these datasets, we conduct a series of experiments to assess model performance across various settings, including table-only input, chart-only input, and combined table–chart input. 
We also evaluate performance across different chart types: basic bar charts, symbol bar charts, line charts, and swapped charts.
Our results from 12 multimodal LLMs show that current models struggle to verify claims when evidence is presented as charts but perform better when the same information is provided in tables, indicating a strong reliance on structured, text-like input. 
This discrepancy highlights a notable limitation in current multimodal LLMs' ability to interpret visual data, even when the content is semantically equivalent.
To better understand this gap, we asked human annotators to complete the same tasks. 
Unlike the models, humans performed consistently well across both formats, suggesting that the observed difficulty stems from model limitations rather than task ambiguity.
These findings underscore the need for improved multimodal reasoning and better alignment between visual and textual representations in future model development.

In summary, our main contributions are as follows.
(1) We extend two existing datasets, SciTabAlign and ChartMimic, to create enhanced versions that support scientific claim verification using different evidence formats representing the same information.
(2) We comprehensively evaluate 12 multimodal LLMs under three input settings: table-only, chart-only, and a combination of both. We also investigate performance across different chart types.
(3) Our experimental results show that current multimodal LLMs struggle to process chart-based input, while performing better with text-based table input. Human performance on the same task confirms that people do not face similar difficulties with different evidence formats, highlighting a limitation in current multimodal LLMs.

% \todo{
% Contributions:
% We are the first to investigate how different input formats affect the results of the claim verification task.
% }

% Although our dataset is small, but we are the first one conduct this kind of investigation, which is important for assisting researcher in review systems ...

\section{Related Work}
\label{sec:related_work}

\begin{table*}[h]
  \centering
   \resizebox{\textwidth}{!}{
    \begin{tabular}{ l l l l l r l l}
\toprule
        \textbf{Year} & \textbf{Name} & \textbf{Main Task} & \textbf{Input} & \textbf{Output} & \textbf{Size}  & \textbf{Domain} & \textbf{Dataset Creation} \\ 
        \midrule
        
        2021 & \makecell[l]{SEM-TAB-FACTs \\ \cite{wang-etal-2021-semeval}} & Verdict & Single table + claim & Label & 5,715 & ScienceDirect & Crowdsourcing \\ 
        
        2023 & \makecell[l]{SciTab \\ \cite{lu-etal-2023-scitab} } & Verdict & Single table* + claim & Label  & 1,225 & Computer science & \makecell[l]{Authentic claims + \\ LLM with human verification} \\ 
        
        2025 & \makecell[l]{SciAtomicBench \\\cite{zhang2025atomicreasoningscientifictable}} & Verdict & Single table + claim & Label & 2,568 & Multi & LLM with human verification \\ 
        % Material science, medical science, finance, and computer science
        
        2025 & \makecell[l]{SciVer \\ \cite{wang-etal-2025-sciver}} & Verdict & \makecell[l]{Multi paragraphs, \\ tables, charts + claim} & Label & 3,000 & Computer science & Experts \\ 
        
        2025 & \makecell[l]{MuSciClaims \\ \cite{lal2025musciclaims}} & Verdict & Figures + claim & Label  & 918 & Life sciences & Authentic claims + experts \\ 
        
        2025 & \makecell[l]{SciTabAlign \\ \cite{ho-etal-2025-table}} & Verdict & Single table* + claim & Label & 372 & Computer science & SciTab + Experts \\

        \midrule
        2020 & \makecell[l]{PlotQA \\ \cite{Methani_2020_WACV}} & QA & Single chart + question & Answer & 28.9M & \makecell[l]{Online data sources  \\ (e.g., Open Government Data)} & Semi-automatic \\ 
        
        2022 & \makecell[l]{ChartQA \\ \cite{masry-etal-2022-chartqa}} & QA & Single chart + question & Answer & 9,608 & \makecell[l]{Various Websites \\ (Statista, Pew, OWID, OECD)}  & Crowdsourcing \\ 
        
        2024 & \makecell[l]{ChartBench \\ \cite{xu2024chartbench}} & QA & Single chart + question & Answer &  18.9K & Kaggle & LLM with human verification \\ 
        
        2024 & \makecell[l]{CharXiv \\ \cite{wang2024charxiv}} & QA & Single chart + question & Answer & 2,323 & arXiv & Humans \\ 
        
        2024 & \makecell[l]{MMC-benchmark \\ \cite{liu-etal-2024-mmc}} & QA & Chart(s) + question & Answer & 2,000 & \makecell[l]{Various sources \\ (e.g., arXiv, Statista, VisText)} & Humans + LLM \\ 
        
        2025 & \makecell[l]{DomainCQA \\ \cite{zhong2025domaincqa}} & QA & Single chart + question & Answer & 1,890 & Astronomy & Semi-automatic \\ 
        
        2025 & \makecell[l]{ChartQAPro \\ \cite{masry-etal-2025-chartqapro}} & QA & Single chart + question & Answer & 1,948 & \makecell[l]{Various Websites \\ (Pew, Tabelau, PPIC, OWID)} & Humans + LLM \\

        \midrule
        2024 & \makecell[l]{SciTabQA \\ \cite{ghosh-etal-2024-robust}} & QA & Single table* + question & Answer & 822 & Computer Science & Humans \\ 
        
        2024 & \makecell[l]{SciTaT \\ \cite{zhang-etal-2025-scitat}} & QA & \makecell[l]{Single table* + text \\ + question} & Answer & 953 & Computer Science & LLM with human verification \\

        \midrule

        2024 & \makecell[l]{MatPlotBench \\ \cite{yang-etal-2024-matplotagent} } & Data2Chart & Raw data + Instruction & Chart & 100 &  Matplotlib, Origin & Crawl \\ 

        2025 & \makecell[l]{Plot2Code \\ \cite{wu-etal-2025-plot2code} }& Chart2Code & Chart(s) + Instruction & Code & 132 & \makecell[l]{Python’s matplotlib, \\ Python’s plotly and R’s plotly} & Crawl \\ 
        
        2025 & \makecell[l]{ChartMimic \\ \cite{yang2025chartmimic}} & Chart2Code & Single chart + Instruction & Code & 4,800 & \makecell[l]{arXiv, Twitter, Reddit, \\ Matplotlib, Stackoverflow} & Humans \\ 
        
        2025 & \makecell[l]{ChartEdit \\ \cite{zhao-etal-2025-chartedit}} & Chart2Code & Single chart + Instruction & Code & 1,405 & arXiv & \makecell[l]{Humans + \\ LLM with human verification} \\

        \bottomrule
    \end{tabular}
    }
  \caption{
  Existing related datasets and their information. It is noted that our list is not comprehensive, as we only include datasets related to experimental results or chart numbers. Also, we focus solely on tasks involving scientific papers; therefore, we may ignore datasets from the Wikipedia domain. 
  Due to space constraints, we use the term \textit{Verdict} to represent the \textit{claim verification} task.
  * indicates that the table provided in the dataset is represented in a text-based table format (e.g., JSON), which can be directly used to generate a chart.
  }
  \label{tab:dataset}
\end{table*}

\subsection{Claim Verification}
Claim verification or fact-checking is a well-established research task, with numerous datasets proposed across various domains~\cite{guo-etal-2022-survey}, such as news articles~\cite{wang-2017-liar} and Wikipedia-based resources~\cite{thorne-etal-2018-fever}. In this work, we focus specifically on datasets that involve tables or charts within scientific papers. Table~\ref{tab:dataset} provides a summary of existing datasets relevant to this domain.
Although datasets such as TabFact~\cite{Chen2020TabFact}, InfoTabs~\cite{gupta-etal-2020-infotabs}, and FEVEROUS~\cite{aly2021feverous} also include tables, they are based on Wikipedia and therefore fall outside the scope of our study. As a result, we do not include them in the table.

An important aspect of scientific claim verification is the authenticity of the claims being evaluated. Among the existing datasets, only SciTab, MuSciClaims, and SciTabAlign reuse authentic claims extracted directly from original scientific papers.
Furthermore, we observe that current datasets for claim verification do not include both table and chart formats representing the same underlying information. While some datasets do contain tables, these are often embedded as figures rather than provided in a structured, text-based format (e.g., JSON). 
This limits the ability to programmatically generate corresponding charts.

\subsection{Other Scientific Tasks Involving Tables and Figures}

In addition to claim verification, several related tasks have been explored in the literature, including question answering (QA), chart-to-code generation, and chart-to-text generation.
For the QA task, the objective is to answer natural language questions based on structured data such as tables or visual data such as charts. 
Existing datasets can be broadly categorized by their input modality: some focus solely on charts (e.g.,~\cite{Methani_2020_WACV, masry-etal-2022-chartqa, xu2024chartbench, wang2024charxiv, zhong2025domaincqa, masry-etal-2025-chartqapro}), others on tables (e.g.,~\cite{ghosh-etal-2024-robust, zhang-etal-2025-scitat}), and a few incorporate both charts and tables (e.g.,~\cite{foroutan-etal-2025-wikimixqa}). These tasks share similarities with claim verification in that they require comprehension of structured or semi-structured data, but differ in terms of output type and the specific reasoning involved.

The chart-to-code generation task, sometimes referred to as derendering, involves generating the underlying code (e.g., Matplotlib) that would produce a given chart. Some studies focus solely on recovering the chart specification from the image~\cite{wu-etal-2025-plot2code, yang2025chartmimic}, while others also address chart editing by predicting modifications or supporting interactive changes to the chart~\cite{zhao-etal-2025-chartedit}.
In contrast, the chart-to-text generation task aims to produce natural language descriptions or summaries of charts. While it shares visual understanding components with chart-to-code generation, the output is unstructured text, which introduces distinct challenges in representation and language generation~\cite{kantharaj-etal-2022-chart}.
Our research differs from both of these directions. Rather than focusing on converting charts to code or to tables (which can be derived from the code), we investigate how different evidence formats such as tables and charts affect model performance on the claim verification task. This shifts the focus from generation to reasoning and evidence interpretation.

\section{Datasets}
In this section, we first introduce the existing datasets, SciTabAlign and ChartMimic, which are derived from scientific papers.
We then describe the process we applied to create enhanced versions, SciTabAlign+ and ChartMimic+, which are used in our experiments.

\subsection{Existing Datasets}

\begin{figure*}[h]
    \centering
    \includegraphics[width=0.81\textwidth]{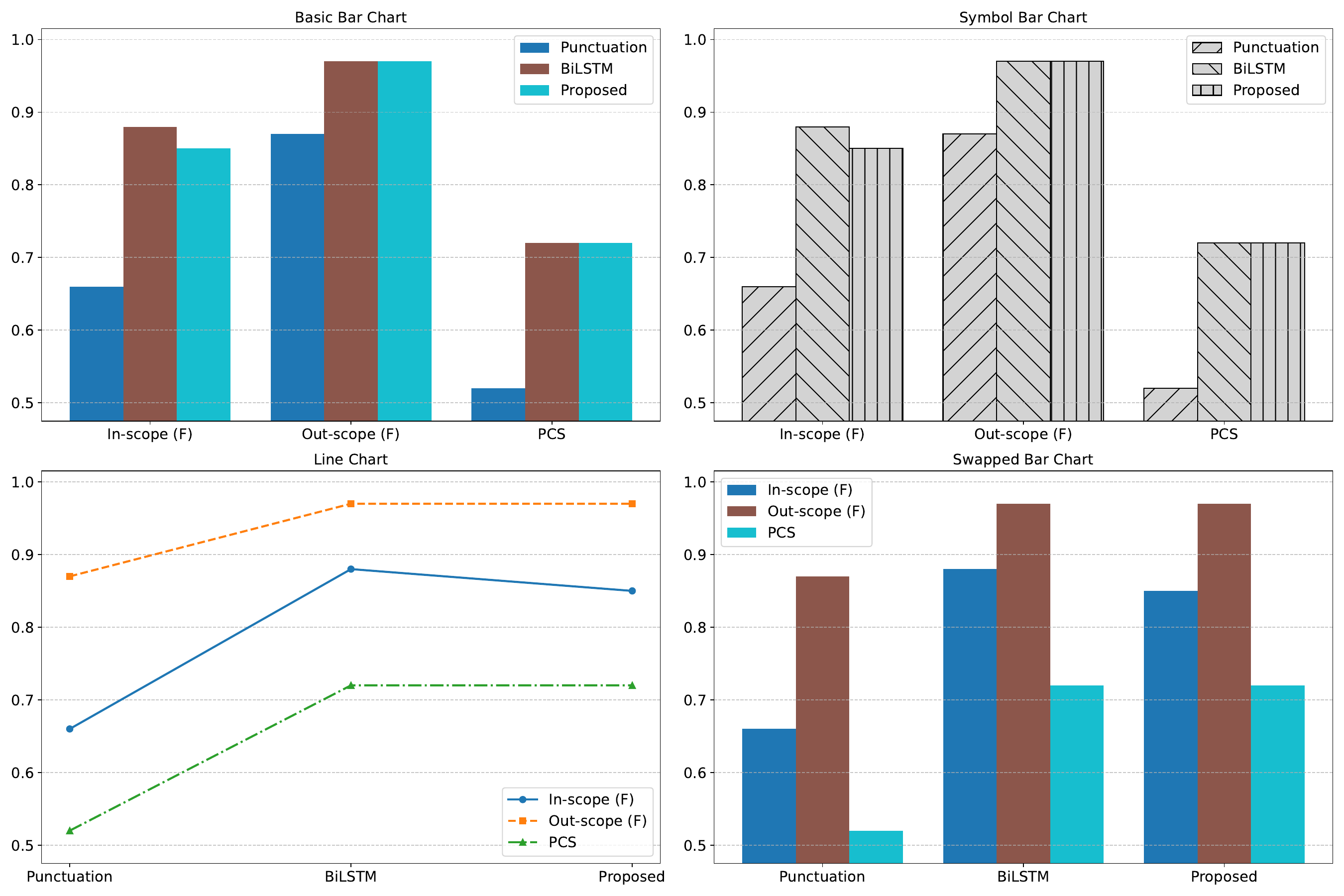}
    \caption{An example of four chart types used in SciTabAlign+.
    }
    \label{fig:example_4types}
\end{figure*}

\paragraph{SciTab and SciTabAlign.}
SciTab~\cite{lu-etal-2023-scitab} is a scientific claim verification dataset that contains authentic scientific claims. 
Its tables are presented in text-based format, making them suitable for generating corresponding charts. SciTabAlign~\cite{ho-etal-2025-table} is an extended version of SciTab, with explanations added for each claim label.
Additionally, ambiguous cases have been removed, particularly those where the available information is insufficient to perform the task or where issues with the table or claim introduce ambiguity.
Since our goal is to investigate the robustness of multimodal LLMs in handling different evidence formats, we chose to work exclusively with SciTabAlign, where all ambiguous cases have been removed.
SciTabAlign includes 136 tables and 372 claims. 
Each sample consists of a text-based table format, a table caption, a claim, and a label for the claim. 
The label can be either Supported or Refuted.

\paragraph{ChartMimic.}
ChartMimic~\cite{yang2025chartmimic} is a specially designed dataset for the chart2code task. 
It consists of two sub-datasets: Direct Mimic and Customized Mimic.
Direct Mimic is intended to evaluate a model’s ability to generate code based on a given chart image.
In contrast, Customized Mimic tests a model’s ability to follow specific instructions to mimic the style of a given chart and use that style to generate code for new data. The resulting code is then used to produce a new chart.
The dataset includes a wide variety of chart types, from popular ones like bar charts and line charts to less common ones like 3D charts.
For the Direct Mimic task, the dataset contains 600 original samples. 
Each sample consists of a PNG image file and a corresponding Python code file that can be used to generate the chart.

% \paragraph{ChartQA.}

% ChartQA \cite{masry-etal-2022-chartqa} is a well-known benchmark for chart-based question answering. 
% The charts in this dataset are sourced from four major platforms: Statista, Pew Research, Our World in Data (OWID), and the Organisation for Economic Co-operation and Development (OECD). 
% The dataset is divided into two subsets: ChartQA-H and ChartQA-M. 
% In ChartQA-H, all question-answer pairs are manually created by human annotators via Amazon Mechanical Turk, while in ChartQA-M, the QA pairs are automatically generated. Both subsets are randomly split into training, validation, and test sets. 
% %
% For our experiments, we use the test set of ChartQA-H, which contains 1,250 human-created QA pairs.
% The dataset includes three main types of charts: bar charts, pie charts, and line charts. 
% Each sample includes a question, an answer, and an image file name, from which the corresponding table-format information can be retrieved from the table pool.
% %
% We use the original version of the ChartQA dataset without applying any additional processing.
% Each chart is provided with a paired data table, which we utilize in our experiments.

\subsection{Datasets Used in Our Experiments}
\label{sec:sub_dataset}
% \todo{Diversity, various difficulty levels, authentic claims/graphs, data leakage}

\paragraph{Processed Version of SciTabAlign (SciTabAlign+).}

We construct SciTabAlign+ through a series of processing steps, as follows.
(1) We first normalize the table data by removing HTML-like tags (e.g., \texttt{<bold>}, \texttt{<italic>}), bracket tags (e.g., [BOLD], [ITALIC]), and standardizing numeric values so that they can be easily converted to floats for chart generation. 
After this process, we retain 70 out of the original 136 tables.
We manually inspected the discarded tables and found several issues. Many contained empty cells, making them difficult to process automatically for accurate chart generation. 
Additionally, some tables included numerical values with ambiguous units (e.g., ``M'' for million), or textual information instead of score values, which are less relevant for our visualization purposes.
As a result, we decided not to generate charts for these tables.
(2) Starting with the 70 selected tables and 162 associated claims, we designed four types of charts:
\textbf{Basic bar charts}, where different colors represent different bars;
\textbf{Symbol bar charts}, where we replace colors with symbols such as ``/'' or ``-'' to form visual patterns;
\textbf{Line charts}, using lines to connect data points; and
\textbf{Swapped charts}, where we interchange the x-axis labels by transforming `methods' into `metrics' and vice versa.
Figure~\ref{fig:example_4types} shows an example of the four chart types used in SciTabAlign+.
We consider this extended dataset a \textit{plus} version of SciTabAlign, called SciTabAlign+, where each claim can now be supported by both table and chart-based evidence.

In summary, our dataset includes 372 claims with table evidence and 648 claims (162 for each of the four chart types) with chart evidence.
To ensure a fair comparison between the table-evidence and chart-evidence formats, if a claim contains references such as ``the results in Table 4 show that...,'' we replace the word table with figure when evaluating on the chart-only setting.

\paragraph{Processed Version of ChartMimic (ChartMimic+).}
We focus only on line charts and bar charts, excluding other types of charts, such as 3D charts, from the scope of this research. 
This choice is motivated by the fact that bar and line charts are not only the most common chart types in the ChartMimic dataset but also the most frequently used for presenting experimental results in academic papers. 
From the Direct Mimic subtask in ChartMimic, which contains 600 charts, we select only bar and line charts. 
Specifically, we obtain 70 line charts and 80 bar charts.
As mentioned above, each sample in ChartMimic consists of a PNG image file and a corresponding Python file used to generate the chart.
Using the provided Python code, we automatically extract the underlying table data.
We then present the following information to the annotators (four NLP researchers): the extracted table, the corresponding bar or line chart, and the original Python code.
Annotators are asked to verify and edit the table so that it accurately reflects the chart.
If the chart and table are deemed suitable for the claim verification task, we ask the annotators to write one supported claim and one refuted claim for each table.
We encourage annotators to write complex claims rather than simple comparative ones such as ``A is better than B.''
During the annotation process, we decided to exclude sub-charts, as our focus is on single-chart analysis. Multi-chart scenarios are left for future work.
We also exclude charts generated using \texttt{np.random.normal} because they do not represent real data suitable for scientific content.

In total, we obtain 152 claims based on 52 bar charts and 24 line charts. 
To ensure consistency between each chart and its corresponding table, we include a caption field where annotators can add any information shown within the chart, such as embedded captions.

\section{Experimental Setup}

\paragraph{Settings.}
We run our experiments under three settings: table-only input, chart-only input, and a combination of both table and chart as input. For SciTabAlign+, we use four different types of charts: bar charts, symbol charts, line charts, and swapped charts.

\paragraph{Models.}
We conduct our investigation using 12 open-source multimodal LLMs from four different families, all capable of processing both image and text inputs. 
Specifically, the models include InternVL3 (1B, 8B, 14B, and 38B)~\cite{zhu2025internvl3exploringadvancedtraining}, Qwen-VL 2.5 (3B, 7B, 32B, and 72B)~\cite{bai2025qwen25vltechnicalreport}, LLaVA-v1.6 (llava-v1.6-mistral-7b, llava-v1.6-vicuna-13b, and llava-v1.6-34b)~\cite{li2024llava}, and Llama-3.2 (11B-Vision)~\cite{grattafiori2024llama3herdmodels}. We use the instruct-tuned versions of models in our experiments.

\paragraph{Promptings.}
Following the previous work~\cite{wang-etal-2025-sciver}, we use zero-shot Chain-of-Thought prompting~\cite{10.5555/3600270.3601883} as the primary evaluation setting in our paper.
Figure~\ref{fig_app_guideline} shows the prompting used in experiments combining chart and table inputs.
When no evidence type is used, we set it to \textit{None} and simplify phrases like ``Use the provided table and image'' to ``Use the provided table.''
% \todo{
% However, to examine the effect of prompt variations on the results, we also provide an analysis in Section~\ref{sec:analyses}.
% }

% Adopted Zeroshot CoT Prompting
% breakable,
\begin{figure}[ht!]
\centering
\begin{tcolorbox}[colback=gray!5, colframe=gray!50, coltitle=black, fonttitle=\bfseries, title=Adopted Zero-shot CoT Prompting,    left=5pt, right=5pt, boxsep=3pt]
\begin{lstlisting}[
  numbers=none,
  xleftmargin=0em,
  framexleftmargin=0em,
  aboveskip=0pt,
  belowskip=0pt,
  basicstyle=\ttfamily\scriptsize\color{black},
  breaklines=true,   
  columns=fullflexible
]
[
  {
    "type": "image",
    "image": "path/to/image"
  },
  {
    "type": "text",
    "text": "{table information} \n Use the provided table and image, predict the label for this claim: {claim}; the label can be Supported or Refuted. Think step by step before answering. Please format your final answer within brackets as follows: <ans> YOUR ANSWER </ans>"
  }
]
\end{lstlisting}

\end{tcolorbox}
\caption{
Input structure used for zero-shot CoT prompting with combined input from both the chart and the table.
}
\label{fig_app_guideline}
\end{figure}

\begin{table*}[ht!]
  \begin{center}
    \resizebox{0.99\textwidth}{!}{% 
    \begin{tabular}{l |c >{\columncolor{gray!30}}c | c c c c >{\columncolor{gray!30}}c | >{\columncolor{gray!30}}   c}
    \toprule
        \textbf{Model} & \textbf{Table (All)} & \textbf{Table (162)} & \textbf{Basic} & \textbf{Symbol} & \textbf{Line} & \textbf{Swapped} & \textbf{Avg.} & \textbf{Chart + Table} \\ \midrule
        
        Qwen2.5-VL-3B & 52.7 & 53.6 & 42.1 & 38.3 & 37.2 & 42.1 & 39.9 & 50.4 \\ 
        
        Qwen2.5-VL-7B & 75.7 & 80.0 & 61.1 & 57.6 & 58.9 & 55.5 & 58.3 & 75.9 \\ 
        
        Qwen2.5-VL-32B & 84.6 & 86.2 & 70.6 & \textbf{67.9} & 65.1 & 66.8 & 67.6 & 86.2 \\ 
        
        Qwen2.5-VL-72B & \textbf{88.5} & \textbf{86.3} & \textbf{70.7} & 63.8 & \textbf{68.8} & \textbf{70.6} & \textbf{68.5} & 88.0 \\ 

        \midrule
        Llama-3.2-11B-Vision & 68.9 & 68.8 & 52.3 & 51.2 & 56.3 & 50.3 & 52.5 & 59.8 \\

        \midrule
        LLaVA-v1.6-Mistral-7B & 52.3 & 57.6 & 57.6 & 56.9 & 55.4 & 60.8 & 57.7 & 58.2 \\ 
        
        LLaVA-v1.6-Vicuna-13B & 48.8 & 49.8 & 43.3 & 36.6 & 36.6 & 35.8 & 38.1 & 48.4 \\ 
        
        LLaVA-v1.6-34B & 60.2 & 56.7 & 32.3 & 32.3 & 35.6 & 33.5 & 33.4 & 37.1 \\ 

        \midrule
        InternVL3-1B & 31.1 & 32.6 & 21.7 & 28.1 & 26.1 & 17.1 & 23.3 & 34.1 \\
        
        InternVL3-8B & 69.9 & 70.4 & 57.3 & 50.8 & 59.8 & 55.0 & 55.7 & 70.2 \\ 
        
        InternVL3-14B & 81.5 & 81.1 & 62.1 & 59.1 & 61.9 & 66.3 & 62.4 & 84.9 \\
        
        InternVL3-38B & 80.7 & 82.4 & 64.7 & 62.6 & 61.1 & 61.4 & 62.5 & \textbf{88.8} \\ 
        \bottomrule
    \end{tabular}
    }
    \caption{
    Macro-F1 scores of the models on the \textbf{SciTabAlign+} dataset under three input settings: table-only, chart-only, and the combination of chart and table. For the combined setting, we use the basic chart type.
    \textbf{Table (All)} refers to results on the full dataset with 372 claims.
    \textbf{Table (162)} refers to results on 162 claims, matching the subset used for each chart type (basic, symbol, line, or swapped) in the chart-only setting.
    \textbf{Avg.} indicates the average score across the four chart types.
    }
    \label{results_scitab}
  \end{center}
\end{table*}

\paragraph{Evaluation.}
Following the evaluation protocols used in the SciTab and SciTabAlign datasets, we adopt macro-F1 as our primary evaluation metric.

\section{Results}

\subsection{SciTabAlign+}
Table~\ref{results_scitab} presents the macro-F1 scores of models on the SciTabAlign+ dataset under three input settings: table-only, chart-only, and the combination of both.

\textbf{[Table vs. Chart]} Comparing table-only and chart-only settings (Table 162 vs. Avg. columns), table-based input consistently outperforms chart-based input across 11 models. 
The only exception is LLaVA-v1.6-Mistral-7B, where performance is nearly identical (57.6 vs. 57.7).
The five largest performance gaps, 23.3, 21.7, 19.9, 18.7, and 18.6, are observed in LLaVA-v1.6-34B, Qwen2.5-VL-7B, InternVL3-38B, InternVL3-14B, and Qwen2.5-VL-32B, respectively.
The remaining gaps range from 9.3 to 17.8.
These substantial gaps indicate that most existing multimodal LLMs struggle with chart inputs while performing better with table inputs.

\textbf{[Table vs. Combination]} 
In some cases (e.g., Qwen2.5-VL-3B, Qwen2.5-VL-7B, Llama-3.2-11B-Vision, and LLaVA-v1.6-34B), using only table input yields better performance than the combined input.
Conversely, some models such as InternVL3-1B, InternVL3-14B, and InternVL3-38B benefit from the combination.
LLaVA-v1.6-34B performs poorly at handling chart information; even when provided with combined input, its performance remains low.
It is worth noting that we used the same prompt and the same setup for all models in our experiments.
We follow the instructions for handling image input as described on each model's Hugging Face page. For example, Qwen recommends using the \textit{process\_vision\_info} function from \textit{qwen\_vl\_utils}, while LLaVA suggests using the \textit{Image} class from \textit{PIL}.

\textbf{[Chart vs. Combination]} 
The combination of chart and table input consistently outperforms chart-only input across all 12 models.
The largest gap is 26.3 on InternVL3-38B, while the smallest is 0.5 on LLaVA-v1.6-Mistral-7B.
For the remaining models, most gaps are larger than 10.0, except for Llama-3.2-11B-Vision (7.3) and LLaVA-v1.6-34B (3.7).
These large gaps between chart-only input and the combination suggest that, although both types of evidence represent the same information, most models still struggle to effectively process chart data.

\textbf{[Comparison Across Different Chart Types]}
Among the four chart types presented in Section~\ref{sec:sub_dataset} (basic bar charts, symbol bar charts, line charts, and swapped charts), we observe that six out of twelve models achieved their best performance on basic bar charts. 
Line charts and swapped charts each had three models achieving the highest scores. Only one model, InternVL3-1B, performed best on the symbol bar chart; however, its score was relatively low, with an macro-F1 of just 28.1.
Notably, one model, Qwen2.5-VL-3B, achieved similar top scores on two chart types, which brings the total count to 13 models instead of 12.
The averages across all 12 models for the four chart types are 53.0, 50.4, 51.9, and 51.3 for basic bar charts, symbol bar charts, line charts, and swapped charts, respectively.
In summary, symbol bar charts appear to be the most challenging, while the other chart types are generally easier for the models, especially basic bar charts, which achieve the highest scores.

\subsection{ChartMimic+}

\begin{table}[ht!]
  \begin{center}
    \resizebox{0.99\columnwidth}{!}{% 
        \begin{tabular}{l c c c }
    \toprule
        \textbf{Model} & \textbf{Table} & \textbf{Chart} & \textbf{Chart + Table} \\

        \midrule
        Qwen2.5-VL-3B & 73.6 & 59.0 & 71.6 \\ 
        Qwen2.5-VL-7B & 83.3 & 87.0 & 87.1 \\ 
        Qwen2.5-VL-32B & \textbf{93.0} & 86.7 & \textbf{95.4} \\ 
        Qwen2.5-VL-72B & 92.1 & \textbf{90.1} & 94.7 \\ 

        \midrule
        Llama-3.2-11B-Vision & 71.5 & 67.6 & 78.9 \\ 

       \midrule
        LLaVA-v1.6-Mistral-7B & 59.8 & 48.1 & 49.7 \\ 
        LLaVA-v1.6-Vicuna-13B & 60.0 & 35.2 & 46.1 \\ 
        LLaVA-v1.6-34B & 61.4 & 33.3 & 42.8 \\ 
        
        \midrule
        InternVL3-1B & 35.2 & 18.0 & 25.0 \\ 
        InternVL3-8B & 90.4 & 80.0 & 82.7 \\ 
        InternVL3-14B & 89.8 & 82.2 & 92.4 \\ 
        InternVL3-38B & 90.1 & 86.2 & 94.7 \\ 

        \bottomrule
    \end{tabular}
    }
    \caption{
    Macro-F1 scores of the models on the \textbf{ChartMimic+} dataset under three input settings: table-only, chart-only, and the combination of chart and table. 
    }
    \label{results_mimic}
  \end{center}
\end{table}

Table~\ref{results_mimic} presents the Macro-F1 scores of the models on the ChartMimic+ dataset under three input settings: table-only, chart-only, and the combination of chart and table.

\textbf{[Table vs. Chart]} When comparing the table-only and chart-only inputs, we observe that 11 models perform better with table input than with chart input.
The four largest performance gaps are 28.1, 24.8, 17.2, and 14.6 for LLaVA-v1.6-34B, LLaVA-v1.6-Vicuna-13B, InternVL3-1B, and Qwen2.5-VL-3B, respectively.
We observe a similar pattern in the SciTabAlign+ dataset, where LLaVA-v1.6-34B also performs well with table input but  worse with chart input.
Qwen2.5-VL-7B is the only model that performs better with chart input than with table input.

\textbf{[Table vs. Combination]} 
For Qwen2.5-VL-3B, LLaVA-v1.6-Mistral-7B, LLaVA-v1.6-Vicuna-13B, LLaVA-v1.6-34B, InternVL3-1B, and InternVL3-8B, the table-only input outperforms the combination of chart and table inputs.
For the remaining models, the combined input performs better than the table-only input.
The four largest gaps, 18.6, 13.9, 10.2, and 10.1, are observed in LLaVA-v1.6-34B, LLaVA-v1.6-Vicuna-13B, InternVL3-1B, and LLaVA-v1.6-Mistral-7B, respectively.
These scores suggest that these models lack the ability to effectively integrate table and chart information in the combination setting.

\textbf{[Chart vs. Combination]} 
Similar to the SciTabAlign+ dataset, we observe that the combination of chart and table inputs outperforms the chart-only input across all 12 models.
This observation further emphasizes that current multimodal LLMs still struggle when working with chart-only information.
Since the chart and table convey the same underlying information, the models should be able to achieve performance comparable to either the table-only or the combined input setting.

% \subsection{ChartQA}

% To test our findings on a larger dataset, we also conduct experiments on the ChartQA dataset.
% It's noted that the task here is question answering, not claim verification.
% ChartQA is the one availabel dataset that we found that have both chart and table available, although their table dataset is not so perfect. 
% The results are presented in Table~\ref{results_chartqa}.

% \begin{table}[ht!]
%   \begin{center}
%     \resizebox{0.99\columnwidth}{!}{% 
%     \input{CameraReady/LaTeX/tables/results_chartqa}
%     }
%     \caption{
%     EM and F1 scores of the models on the ChartQA dataset under three settings: table-only, chart-only, and a combination of chart and table.
%     }
%     \label{results_chartqa}
%   \end{center}
% \end{table}

\subsection{Compared to Humans}
To better understand the difficulty of the task for humans and to test how different evidence formats affect human performance, we randomly selected 50 samples from SciTabAlign+. These samples were split into two sub-tasks, similar to how they are used for models: one using table-only evidence and the other using chart-only evidence.
We presented the two sub-tasks to two different annotators (both Master's students in Computer Science) and asked them to perform the task. 

The resulting Macro-F1 scores are 94.0 for the table-only evidence and 96.0 for the chart-only evidence.
The Pearson correlation between the annotations of the two annotators is 0.887.
These scores indicate that humans perform well with either tables or charts, as the format does not affect their performance when the input information is the same.

\section{Correlation between Tables and Charts}
\label{sec:analyses}

To better understand the behavior of models when using table-only versus chart-only evidence, we analyze the correlation scores between these two settings.
Table~\ref{results_correlation} presents the Pearson correlation scores for the two datasets: SciTabAlign+ and ChartMimic+.

For SciTabAlign+, we observe that the correlation between the table-only and chart-only settings is generally low, with some small models such as InternVL3-1B even showing negative correlation.
Overall, correlation scores tend to be higher for larger models and lower for smaller models.
In contrast, the correlation scores for the ChartMimic+ dataset are higher than those for the SciTabAlign+ dataset.
Large models such as Qwen2.5-VL-32B, Qwen2.5-VL-72B, and InternVL3-38B achieve scores above 0.7, indicating a strong correlation between the two settings.

Considering both datasets, we observe that models from the LLaVA and Llama families show low correlation between table-only and chart-only inputs when solving the task.
For the Qwen and InternVL3 families, smaller models with fewer than 8 billion parameters do not show strong correlation between the two evidence formats.
However, larger models (with more than 32 billion parameters) exhibit strong correlation on ChartMimic+, but not on SciTabAlign+.

\begin{table}[ht!]
  \begin{center}
    \resizebox{0.99\columnwidth}{!}{% 
        \begin{tabular}{l | r r r r | r}
    \toprule
        \textbf{Model} & \textbf{Basic} & \textbf{Symbol} & \textbf{Line} & \textbf{Swap} & \textbf{Mimic+} \\ \midrule
        
        Qwen2.5-VL-3B & 0.132 & 0.143 & 0.132 & 0.132 & 0.295 \\ 
        
        Qwen2.5-VL-7B & 0.246 & 0.237 & 0.203 & 0.178 & 0.657 \\
        
        Qwen2.5-VL-32B & \textbf{0.417} & \textbf{0.301} & 0.341 & 0.352 & 0.710 \\
        
        Qwen2.5-VL-72B & 0.337 & 0.264 & \textbf{0.396} & \textbf{0.446} & \textbf{0.805} \\

        \midrule
        
        Llama-3.2-11B & 0.144 & 0.113 & 0.121 & 0.065 & 0.400 \\

        \midrule
        LLaVA-Mistral-7B & 0.029 & 0.175 & 0.051 & 0.145 & 0.146 \\
        LLaVA-Vicuna-13B & 0.106 & 0.051 & 0.080 & 0.118 & 0.149 \\
        LLaVA-34B & 0.059 & 0.059 & 0.252 & 0.125 & NaN \\

        \midrule
        InternVL3-1B & -0.055 & -0.042 & -0.100 & 0.176 & -0.056 \\
        InternVL3-8B & 0.163 & 0.143 & 0.182 & -0.026 & 0.553 \\
        InternVL3-14B & 0.264 & 0.194 & 0.151 & 0.302 & 0.593 \\
        InternVL3-38B & 0.225 & 0.351 & 0.229 & 0.133 & 0.762 \\

        \bottomrule
    \end{tabular}
    }
    \caption{
    Correlation between using table-only and chart-only evidence for the two datasets: SciTabAlign+ and ChartMimic+.
    For SciTabAlign+, we include four different chart types, which are presented as four columns: Basic Chart, Symbol Chart, Line Chart, and Swapped Chart.
    }
    \label{results_correlation}
  \end{center}
\end{table}

\section{Conclusion}

In this paper, we extend two existing datasets, SciTabAlign and ChartMimic, to create enhanced versions that support scientific claim verification across different evidence formats conveying the same underlying information.
Using these datasets, we conduct a comprehensive evaluation of 12 multimodal LLMs under three input settings: table-only, chart-only, and a combination of both. We also analyze model performance across different chart types.
Our experimental results reveal that current multimodal LLMs perform better with table-based inputs but struggle with chart-based evidence. 
In contrast, human performance remains consistent across both formats, suggesting that the observed discrepancies reflect limitations in the models rather than in the task design.
To improve the reliability of automated scientific review systems, future research should focus on enhancing multimodal LLMs’ ability to interpret and reason over diverse visual formats, particularly charts, as a crucial step toward robust scientific claim verification.

% \section{Acknowledgments}

% \section*{Limitations}
% Our work has several limitations ...

\section*{Acknowledgments}
We would like to thank the anonymous reviewers for their feedback and suggestions on the improvement of the paper.
We also thank Tian Cheng Xia and Truc Hoang for their help with the human annotation tasks.
We are grateful to the National Institute of Informatics (NII) for supporting this research.
This work was supported by JSPS KAKENHI Grant Number 24K03231.

% \bigskip
% \noindent Thank you for reading these instructions carefully. We look forward to receiving your electronic files!

\bibliography{aaai2026}

\appendix
\section{Experimental Details}
\paragraph{Running.}
We run all models on either a single NVIDIA A100 80 GB GPU or two of them.
We use max\_new\_tokens=1024 for all models.

\paragraph{Evaluation.}
We adopt the precision\_recall\_fscore\_support function from scikit-learn and use it in our evaluation.

\end{document}